%% file: main.tex
\documentclass[sigconf,screen]{acmart}

\startPage{1}
\setcopyright{none}

\usepackage{amsmath,amsfonts,bbm}
\usepackage{graphicx}
\usepackage{textcomp}
\usepackage{xcolor}
\usepackage{fancyhdr}
\usepackage{enumitem}
\usepackage{hyperref}
\usepackage{rotating}
\usepackage{booktabs}
\usepackage{balance}
\usepackage{algorithm}
\usepackage[noend]{algpseudocode}
\usepackage{multirow}
\usepackage{soul}
\usepackage{blkarray}
\usepackage{lipsum,wrapfig}
\usepackage{caption}
\usepackage{subcaption}
\usepackage{tikz}
\usepackage{cleveref}

\definecolor{myblue}{HTML}{4682B4}

\hypersetup{
	colorlinks=true,
	linkcolor = blue,
	citecolor=myblue}

\setlist{noitemsep, leftmargin=*, topsep=0pt, partopsep=0pt}

\crefformat{section}{\S#2#1#3} 
\crefformat{subsection}{\S#2#1#3}
\crefformat{subsubsection}{\S#2#1#3}

\newcommand{\SYSTEM}{Cello}

\newcommand{\LUP}{Latency-Under-Power}
\newcommand{\EUL}{Energy-Under-Latency}

\newcommand{\RS}{RS}
\newcommand{\BO}{BO}
\newcommand{\BOST}{BO-ST}
\newcommand{\BOTC}{BO-TC}
\newcommand{\BOIM}{BO-IM}
\newcommand{\BONN}{BO-NN}
\newcommand{\BOGB}{BO-GB}
\newcommand{\Bliss}{Bliss}

\newcommand{\x}{{\mathbf{x}}}

\DeclareMathOperator*{\argmax}{arg\,max}
\DeclareMathOperator*{\argmin}{arg\,min}

\newcommand*\circled[1]{\tikz[baseline=(char.base)]{
		\node[shape=circle,draw,inner sep=0.7pt] (char) {#1};}}

\newcommand*\rectangled[1]{\tikz[baseline=(char.base)]{
		\node[shape=rectangle,draw,inner sep=1.5pt] (char) {#1};}}

\begin{document}

\title{\SYSTEM{}: Efficient Computer Systems Optimization with Predictive Early Termination and Censored Regression}

\author{Yi Ding}
\affiliation{
	\institution{MIT CSAIL}
	\city{Cambridge, MA}
	\country{USA}}
\email{ding1@csail.mit.edu}

\author{Alex Renda}
\affiliation{
	\institution{MIT CSAIL}
	\city{Cambridge, MA}
	\country{USA}}
\email{renda@csail.mit.edu}

\author{Ahsan Pervaiz}
\affiliation{
	\institution{University of Chicago}
	\city{Chicago, IL}
	\country{USA}}
\email{ahsanp@uchicago.edu}

\author{Michael Carbin}
\affiliation{
	\institution{MIT CSAIL}
	\city{Cambridge, MA}
	\country{USA}}
\email{mcarbin@csail.mit.edu}

\author{Henry Hoffmann}
\affiliation{
	\institution{University of Chicago}
	\city{Chicago, IL}
	\country{USA}}
\email{hankhoffmann@cs.uchicago.edu}

\input{tex/abstract}

\maketitle

\input{tex/intro}
\input{tex/related}

\input{tex/framework}
\input{tex/evaluation}

\input{tex/conclusion}


\bibliographystyle{unsrtnat}
\bibliography{reference}

\end{document}

%% file: tex/abstract.tex
\begin{abstract}

  Sample-efficient machine learning (SEML) has been widely applied to
  find optimal latency and power tradeoffs for configurable computer
  systems.  Instead of randomly sampling from the configuration space,
  SEML reduces the search cost by dramatically reducing the number of
  configurations that must be sampled to optimize system goals (e.g.,
  low latency or energy).  Nevertheless, SEML only reduces one
  component of cost---the total number of samples collected---but does
  not decrease the cost of collecting each sample.  Critically, not
  all samples are equal; some take much longer to collect because they
  correspond to slow system configurations. This paper present
  \SYSTEM{}, a computer systems optimization framework that reduces
  sample collection costs---especially those that come from the
  slowest configurations. The key insight is to predict ahead of time
  whether samples will have poor system behavior (e.g., long latency
  or high energy) and terminate these samples early before their
  measured system behavior surpasses the termination threshold, which
  we call it \emph{predictive early termination}. To predict the
  future system behavior accurately before it manifests as high
  runtime or energy, \SYSTEM{} uses censored regression to produces
  accurate predictions for running samples. We evaluate \SYSTEM{} by
  optimizing latency and energy for Apache Spark workloads. We give
  \SYSTEM{} a fixed amount of time to search a combined space of
  hardware and software configuration parameters. Our evaluation shows that compared to the state-of-the-art SEML approach in computer systems optimization, \SYSTEM{} improves latency by 1.19$\times$ for minimizing latency under a power constraint, and improves energy by 1.18$\times$ for minimizing energy under a latency constraint.

\end{abstract}

%% file: tex/intro.tex
\section{Introduction}

Optimizing latency and power tradeoffs (e.g., minimizing latency while
meeting a power constraint) is crucial to modern computer systems. To
achieve these goals, hardware and software configuration
parameters---e.g., number of cores, core frequency, memory management,
etc.---are exposed to users such that both the underlying system and
application can be configured to operate at an optimal point in the
latency-power tradeoff
space~\cite{xu2015hey,colin2018reconfigurable,blocher2021switches}.
Selecting an optimal configuration, however, is a challenging problem
as the size and complexity of the configuration space require
intelligent methods to avoid local optima \cite{LEO,CALOREE}.

\textbf{Sample Efficiency.} Machine learning techniques have proven
effective solutions to optimizing computer system
configurations~\cite{herodotou2020survey,penney2019survey,ZhangH2019learned}.
While a variety of such methods have been applied, they require a
large sampling effort; i.e., generating a candidate configuration,
running it to measure its high-level \emph{system behavior} (e.g.,
latency and energy), and then building a model to predict the system
behavior of unsampled configurations.  Among these techniques, an
appealing direction is sample-efficient machine learning (SEML), which
reduces the number of samples required to find the optimal
configuration
~\cite{ipek2006efficiently,alipourfar2017cherrypick,ding2019generative,nardi2019practical,patel2020clite}.
A powerful example of SEML is Bayesian
optimization~\cite{alipourfar2017cherrypick,nardi2019practical,patel2020clite}.
Bayesian optimization intelligently selects samples that provide the
most information for the model updates, and this intelligence reduces
the number of samples required to find the configuration that produces
optimal systems behavior.  Besides Bayesian optimization, other
sample-efficient optimizers have been proposed for specific systems
problems~\cite{ipek2006efficiently,petrica2013flicker,venkataraman2016ernest,ding2019generative}.

\textbf{Cost of Samples.} SEML reduces the cost for computer systems
optimization by reducing the number of samples to find the optimal
configuration.  There is, however, a significant additional cost: the
time to collect each sample. Critically in computer systems
optimization, the samples corresponding to the worst configurations
take a much longer time to collect than others. For instance, a slow
configuration can take 10$\times$ the execution time of the optimal
configuration on the same Apache Spark workload~\cite{yu2018datasize}.
Thus, there is a need for reducing the cost of individual sample
collection and not just the total number of samples.

\textbf{Compute Efficiency}. We note that achieving \emph{compute
	efficiency} for SEML should reduce both the number of samples and
the cost of collecting each sample. To reduce the cost of collecting
each sample,
prior work replies on \emph{measured early termination} that
terminates samples when the measured system behavior surpasses a given
\emph{termination threshold} (e.g., a target
latency)~\cite{hutter2013bayesian,eggensperger2020neural}; and this
threshold typically distinguishes between good and poor system
behavior. For instance, if the termination threshold is 40 seconds and
the true latency of the workload running at a particular configuration is 60
seconds, measured early termination techniques will terminate samples
when their runtime surpasses 40 seconds. However, this approach still
pays the cost of waiting for the sample to hit the termination
threshold, which is compute-inefficient.

\subsection{Compute Efficiency with \SYSTEM{}}

In contrast to prior work based on measured early termination, we
introduce \emph{predictive early termination}.  This approach monitors
running samples, predicts whether the sample will surpass the
threshold, and terminates such unpromising samples before they are
measured to be poor (e.g., long latency or high energy). 

To apply this
insight, we present \SYSTEM{}, an efficient computer systems
optimization framework that augments Bayesian optimization with
predictive early termination. To enable predictive early termination,
\SYSTEM{} must predict, both early and accurately, whether a running
sample will perform poorly; i.e., eventually exceed the termination
threshold, which \SYSTEM{} sets as the best measured behavior.
To do so, \SYSTEM{} applies \emph{censored regression}, a prediction
approach that is designed to model missing data (i.e., samples
that are measured but not finish running) more accurately than techniques that do not
account for the missing
data~\cite{miller1982regression,portnoy2003censored}.  Formally,
\emph{censoring} is a condition where a measurement is only
partially observable due to a given censoring
threshold~\cite{powell1986censored}. Instead of observing the
true measurement, in censored regression we only know that it exceeded the
censoring threshold. For example, if the censoring threshold is 30
seconds, the true latency of a sample that keeps running after 30
seconds is not observed at this moment because it is censored.
Although its true latency is unknown at this moment, we know that it must be at least
30 seconds. The intuition behind censored regression is to model the
censored and uncensored data (in our case, the samples that are still running and finished running before the termination threshold) differently, and
then combine these models in a way that produces more accurate
predictions than standard regression that does not account
for censoring (\cref{sec:res-censor}). 

\subsection{Summary of Results}

We implement \SYSTEM{} and test it on twelve Apache Spark workloads
from HiBench~\cite{huang2010hibench}. We compare \SYSTEM{} to
state-of-the-art Bayesian optimization techniques and different
variants of early termination (including both measured and
predictive). We give each method a fixed time to search a combined
space of hardware and software configuration parameters and find
solutions for two systems optimization problems: (1) minimizing
latency under a power constraint (\LUP{}) and (2) minimizing energy
under a latency constraint (\EUL{}). Our evaluation shows a complete
breakdown of results for different constraints and workloads for
\SYSTEM{} and prior approaches to SEML.  Compared to the prior work,
we find that on average \SYSTEM{} (\cref{sec:res-good}):
\begin{itemize}
\item improves latency by 1.19$\times$ for \LUP{} and energy by 1.18$\times$ for \EUL{} compared to \Bliss{}~\cite{roy2021bliss}, the state of the art SEML approach for computer systems optimizations;
\item improves latency by 1.24--1.61$\times$ for \LUP{} and energy by 1.24--1.34$\times$ for \EUL{} compared to other SEML approaches that use measured early termination;
\item improves latency by  1.90$\times$ for \LUP{} and energy by 2.35$\times$ for \EUL{} compared to the SEML approach that uses predictive early termination based on standard regression, rather than \SYSTEM{}'s censored regression.
\end{itemize}

\subsection{Contributions}

This paper presents the following contributions:
\begin{itemize}
\item Introducing predictive early termination, based on
  predicted---rather than measured---behavior which allows SEML to
  terminate samples much earlier than prior work.
\item Introducing censored regression to predict the system behavior
  of samples and enable predictive early termination. To the best of
  our knowledge this is the first demonstration of the benefits of
  censored regression for computer systems optimization.
\item Presenting \SYSTEM{}, an efficient computer system
  optimization framework that combines predictive early termination
  and censored regression to reduce the cost of applying
  machine learning methods for systems optimization while still
  improving systems outcomes.

\end{itemize}
Incomplete data are often prevalent in computer systems optimization.
To the best of our knowledge, \SYSTEM{} is the first
framework that reduces the time cost per sample by intelligently
taking advantages of incomplete data via censored regression and
predictive early termination. Our work provides a foundation on which
the computer systems community can build new efficient
frameworks that process incomplete data in computer
systems research.

%% file: tex/related.tex
\section{Background and Related Work}\label{sec:related} 

This section discusses related work on machine learning for
configuration optimization (\cref{sec:seml}), early termination
(\cref{sec:early-termination}), and machine learning for prediction (\cref{sec:ml-pred}).

\subsection{Computer Systems Configuration Optimization}\label{sec:seml}

Modern computer systems are increasingly
configurable~\cite{vishwanath2010characterizing,xu2015hey}.  These
configuration parameters have a large effect on high-level systems
behavior including throughput~\cite{belay2014ix},
latency~\cite{van2017automatic}, and energy~\cite{LEO,ding2019generative}. Furthermore, the number of
parameters and their complex interactions create a large search space
with many local optima~\cite{deng2017memory,CALOREE}, which present
challenges for heuristics~\cite{LEO,CALOREE}. To optimize systems
behavior, machine learning techniques have been applied to search this
configuration space for high-performing configurations.

Sample-efficient machine learning (SEML) represents a class of machine
learning techniques that reduce the number of samples required to find
the optimal configuration~\cite{settles2011theories}. SEML approaches
work iteratively, using their current predictions to determine which
new configuration to sample and then update their model based on the
observed behavior of that sample. 
Bayesian optimization is a typical SEML approach that has been applied
in various domains~\cite{frazier2018tutorial}. For example, Cherrypick
uses Bayesian optimization to find optimal cloud
configurations~\cite{alipourfar2017cherrypick}. CLITE uses Bayesian
optimization to schedule workloads for
datacenters~\cite{patel2020clite}. HyperMapper applies Bayesian
optimization to tune compilers~\cite{nardi2019practical}. Bliss uses
Bayesian optimization to tune parallel applications on large-scale
systems~\cite{roy2021bliss}.  Additional works propose
problem-specific SEML approaches, including multi-phase
sampling~\cite{ding2019generative}, optimal experiment design in
Ernest~\cite{venkataraman2016ernest}, and fractional factorial design
in Flicker~\cite{petrica2013flicker}.

\begin{figure}[!htb]
	\centering
			\includegraphics[width=\linewidth]{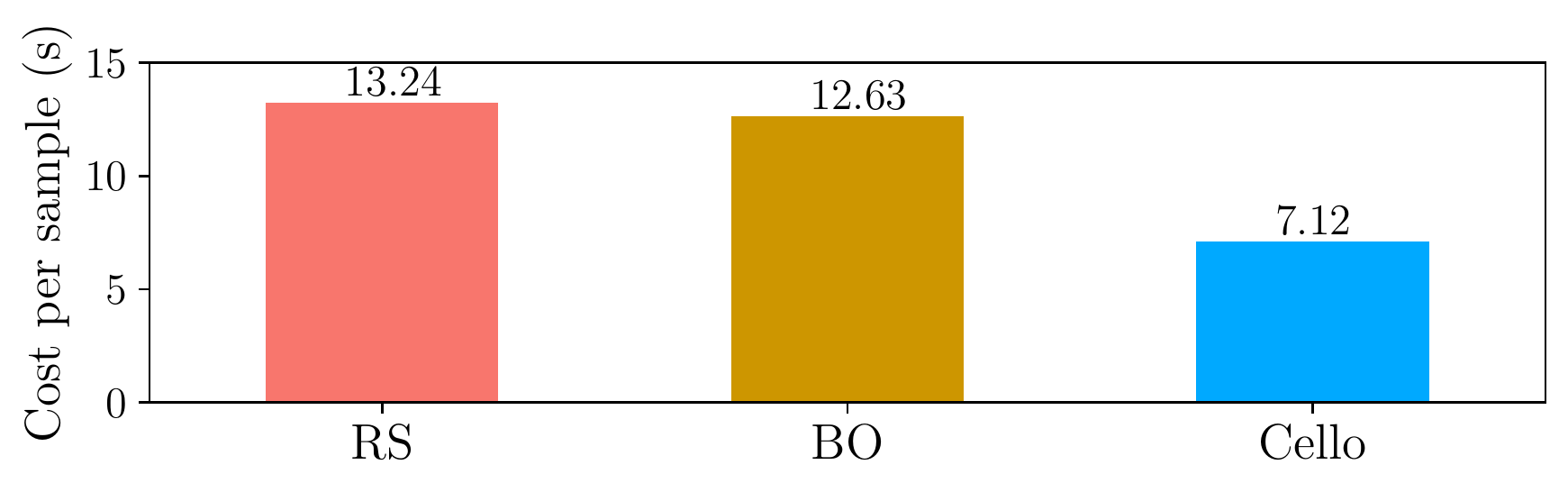}
\caption{Comparing cost per sample between random sampling (RS), Bayesian optimization (BO), and \SYSTEM{} on the \texttt{als} workload.}
\label{fig:motive_cso}
\end{figure}

\textbf{An example of cost per sample.} The SEML approaches above
reduce the cost of computer systems optimization by reducing the
number of samples. However, there is also the opportunity to reduce
the cost of each sample itself. To demonstrate this, we compare the
number of samples and cost per sample between random sampling (not a
SEML approach), Bayesian optimization, and \SYSTEM{}. We run the
Apache Spark workload \texttt{als} from HiBench on a cloud computing
system (details in \cref{sec:hw-systems}) with the goal of finding the
configuration with the lowest latency. 
Figure~\ref{fig:motive_cso} compares the time cost per sample between random sampling, Bayesian optimization, and \SYSTEM{}.
We observe that the time costs per sample from random sampling and Bayesian optimization
approaches are roughly the same, which validates that
Bayesian optimization does not reduce the time cost of collecting samples
\SYSTEM{}, however, focuses on reducing the
cost of each sample. For achieving the same optimal latency,
\SYSTEM{} only needs 7.12s per sample, which is a 44\% reduction in
the time cost per sample compared to Bayesian optimization.

\subsection{Early Termination}\label{sec:early-termination}

\emph{Early termination} aborts poor performing configurations before
they finish, leaving time and resources to sample more promising
configurations.  This approach has been applied to speed up
hyperparameter optimization and deep neural network (DNN) training.
For hyperparameter optimization, HyperBand~\cite{li2017hyperband} and
HyperDrive~\cite{rasley2017hyperdrive} randomly generate a large
number of configurations at one time rather than intelligently
sampling a small number of configurations iteratively, which is much
more compute-intensive than SEML approaches.  For DNN training, prior
work predicts future behavior using learning curves---i.e., the
learning accuracy of an algorithm as a function of its training
time~\cite{swersky2014freeze,domhan2015speeding,Klein2017LearningCP}.
However, Bayesian optimization does not have an analogous structure to
learning curves, so we need a different mechanism for predicting
future behavior and in this work we demonstrate the use of censored
regression for this prediction problem.

For Bayesian optimization, \citet{hutter2013bayesian} and \citet{eggensperger2020neural} apply measured early termination to terminate samples when their measured behavior reaches the termination threshold. \SYSTEM{}, however, uses predictive early termination to terminate samples when their predicted behavior---obtained from censored regression---reaches the termination threshold. This prediction step enables much earlier termination and thus saves more time for \SYSTEM{} to explore more samples compared to prior work (\cref{sec:res-good},\cref{sec:res-samples}).

\subsection{Machine Learning for Behavior Prediction}\label{sec:ml-pred}

A large body of machine learning techniques have been applied to
predict high-level system behavior for resource
management
\cite{ponomarev2001reducing,choi2006learning,lee2007methods,lee2008cpr,martinez2009dynamic,snowdon2009koala,cochran2011pack,zhang2012flexible,sridharan2013holistic,petrica2013flicker,oliner2013carat,Reddi2013,LEO,CALOREE,ding2019generative,zhang2020sinan,shi2019applying,shi2021hierarchical}
and
scheduling~\cite{tesauro2007reinforcement,winter2010scalable,van2012scheduling,quasar,paragon}.
The general strategy is to use low-level, readily-available metrics
(e.g., branch miss rates, IPC) to predict high-level behavior (e.g.,
throughput or
latency)~\cite{lee2006accurate,Ipek2008,bitirgen2008coordinated,snowdon2009koala,dubach2010predictive,Reddi2013,teran2016perceptron,peled2019neural,ding2019generative,bhatia2019perceptron,garza2019bit,bera2021pythia,peled2015semantic}.
For example, \citet{lee2006accurate,lee2010applied} apply linear
regression to predict performance and power. \citet{paragon,quasar} use collaborative filtering to predict job performance for datacenter workloads. \citet{LEO,CALOREE} use hierarchical Bayesian models to predict latency and energy. \citet{garza2019bit} and \citet{bhatia2019perceptron} use perceptrons to predict branch and prefetcher behavior.

\begin{figure}[!htb]
	\centering
	\includegraphics[width=\linewidth]{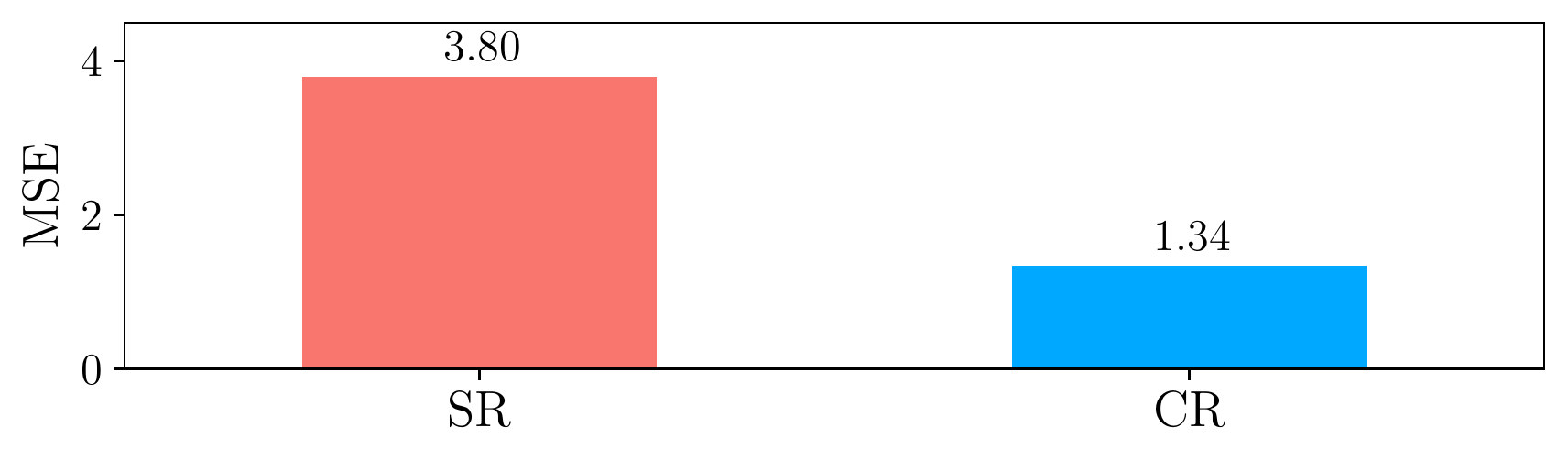}
	\caption{Comparing prediction accuracy between standard regression
		(SR) and censored regression (CR) in terms of mean squared error (MSE) on \texttt{als}.}
	\label{fig:motive_srcr}
\end{figure}

\begin{figure*}[!htb]
	\centering
	\includegraphics[width=0.8\textwidth]{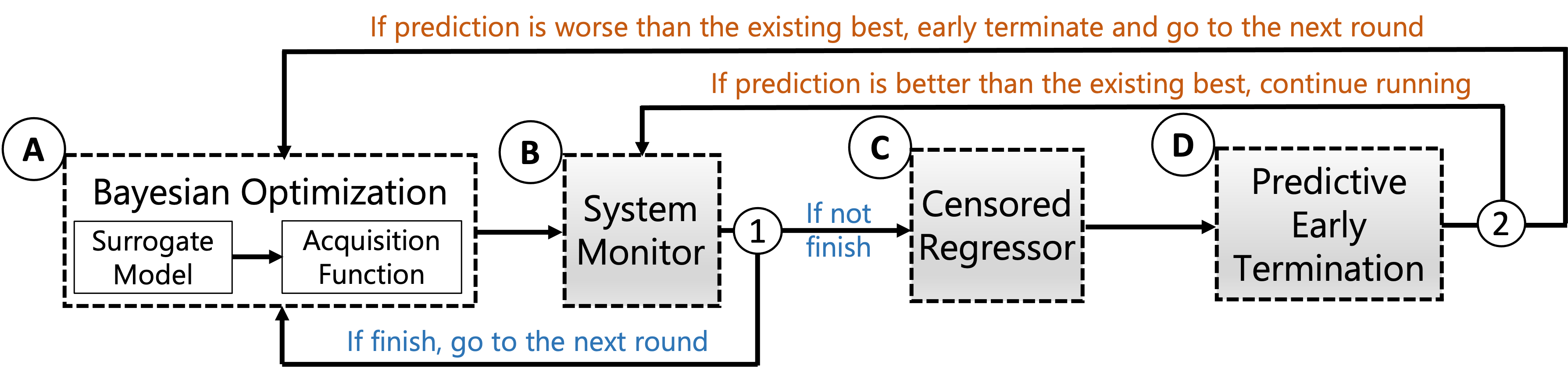}
	\caption{\SYSTEM{} workflow. White Box A is Bayesian
		optimization, and shaded Boxes B to D are unique to \SYSTEM{}. }
	\label{fig:workflow}
\end{figure*}

Most prior work achieves high prediction accuracy relying on
a training set that includes samples in a wide range of behavior.
Bayesian optimization, however, accumulates samples as more iterations finish. Thus, there are
insufficient samples as training data in the early stage of Bayesian
optimization, and these samples are likely to lie in the range of poor
values~\cite{rasley2017hyperdrive}.  Directly training models on these
samples will predict running samples to be poor, which leads to 
a increasing number of predictions of poor values and thus
inaccurate sample selection for the next round~\cite{roy2021bliss}. To
overcome this inaccuracy, \SYSTEM{} utilizes knowledge of the current
elapsed latency and energy at each time interval to generate more
accurate predictions through censored regression.

\textbf{An example of prediction accuracy.} To demonstrate the
benefits of censored regression, we run Bayesian optimization on the
\texttt{als} workload and predict each new configuration's latency
with different predictors: standard and censored regression. We use
gradient boosting trees as our standard regression model due to its
high predictive accuracy~\cite{chen2016xgboost}.
We monitor the workload execution and predict its latency at each
  time interval (see~\cref{sec:eval-method}). The censoring
  threshold for censored regression is the elapsed latency
  observed at each time interval. We compute the mean squared error
  (MSE) between predicted latency and true latency.
  Figure~\ref{fig:motive_srcr} shows the results, where the y-axis
  represents the average MSE over all time intervals, lower is
  better. Censored regression has 2.84$\times$ lower MSE than standard
regression. This example illustrates that censored regression is
effective at improving prediction accuracy when training on samples
collected from Bayesian optimization. The accurate prediction from
censored regression is fundamental to predictive early termination in \SYSTEM{},
as we describe next.

%% file: tex/framework.tex
\section{\SYSTEM{} Design}\label{sec:framework}

\SYSTEM{} is an efficient computer systems optimization
framework that reduces the cost of samples by predictive early termination which
we enable with censored regression. Figure~\ref{fig:workflow}
illustrates the \SYSTEM{} design. After a configuration is selected
from Bayesian optimization (\circled{A}), \SYSTEM{} starts monitoring
it (\circled{B}), and records the current latency and energy at
regular time intervals. At each interval, \SYSTEM{} checks if the
configuration has finished running (Branch \circled{1}). If so,
\SYSTEM{} goes back to Bayesian Optimization (\circled{A}) for the
next round.  Otherwise, \SYSTEM{} predicts the final latency or energy
using the Censored Regressor (\circled{C}).  Then, \SYSTEM{} sends the predictions to Predictive Early Termination (\circled{D}) to determine whether the configuration should be terminated based on
its prediction (Branch \circled{2}). If the prediction is worse than
the existing best from the collected samples, \SYSTEM{} terminates
this sample collection and goes to the next round.
Otherwise, the configuration continues running until the next time
interval. When the configuration finishes, \SYSTEM{} records the data
for this sample and goes to the next Bayesian optimization round
(\circled{A}).

The remainder of this section first provides a brief set of definitions to establish the core concepts, and then describe \SYSTEM{}'s design, and concludes with the \SYSTEM{}'s application to computer systems optimization problems.

\subsection{Definitions}

\SYSTEM{}'s input includes a workload and a list of configuration
parameters over which to optimize. The goal is to find a configuration
that meets the goal of systems optimization---e.g., the configuration
that minimizes latency under a power constraint.  We assume that
\SYSTEM{} can measure high-level systems behavior including latency, power, 
and energy at different time intervals, which is reasonable since many
software frameworks (e.g., Apache Spark~\cite{spark},
MapReduce~\cite{mapreduce}, Cassandra~\cite{cassandra},
HBase~\cite{hbase}, etc) already have this capability as they log
systems behavior at regular time intervals.
\begin{description}
\item[Workload.] A program that runs on a computer system.
\item[Configuration.] A configuration $\x_i$ is a $p$-dimensional vector
  where $p$ is the number of configuration parameters:
\begin{align}\label{eq:vector}
\x_i = [x_{i1},x_{i2},\cdots, x_{ip}],
\end{align}
As SEML sequentially explores configurations, we use index $i$ to
refer to the sequence of configurations explored; $\x_i$ is the $i$-th
configuration.  $x_{ij}$ refers to an individual parameter setting,
the $j$-th parameter, $j\in[p]$. A configuration parameter can either be a hardware feature such as the number of cores, core frequency, etc, or a software feature such as execution behavior, networking, scheduling, etc. A complete list of configuration parameters studied in this paper is in Table~\ref{tbl:conf}.
\item[System behavior.] The high-level system behavior that \SYSTEM{} monitors and optimizes. In this paper, we use latency (i.e., execution
time of a workload), power, and energy consumption as system behavior,
  depending on the specific problem.
\item[Sample.] Given the above, a sample is the pair of a
  configuration along with its system behavior.
\item[Termination threshold.] The termination threshold is a latency
  or energy value that is used to determine whether to
  terminate a sample early or not. Unlike prior approaches~\cite{hutter2013bayesian,eggensperger2020neural}, \SYSTEM{} dynamically updates this threshold based on the best latency or energy seen so far. 
\end{description}
Next, we will describe each labeled box in Figure~\ref{fig:workflow}.

\subsection{Bayesian Optimization}

\SYSTEM{} uses Bayesian optimization (\circled{A}) to select the
configuration at each round~\cite{settles2011theories}. 
Formally,
\begin{align}
\x^\star = \argmin_{\x\in \mathbb{R}^p} f(\x),
\end{align}
where $f(\cdot)$ is the unknown underlying function which is being optimized. 
There are two main components in Bayesian optimization:
the surrogate model and the acquisition function. The surrogate
model learns the unknown underlying function $f(\cdot)$. The
acquisition function selects the most informative
configuration at each round. We describe these two components as
follows.

\subsubsection{Surrogate Model}

Bayesian optimization requires a surrogate model that produces
uncertainty estimates, and \SYSTEM{} requires the surrogate model that
can handle different variable modalities (continuous, discrete, etc.) that exist in configuration space. In this paper, we use random forest as our surrogate model as it satisfies both of these
requirements~\cite{hutter2011sequential,nardi2019practical}. As an ensemble learning method, random forest constructs a multitude of decision trees, where its predictions are the average of the output from its component trees. The uncertainty, therefore, can be obtained by taking the standard deviation from the predictions of each tree~\cite{breiman2001random}. 
For configuration $\x_i$, we denote $\mu_i(\cdot)$ and $\sigma_i(\cdot)$ as its predictive mean and
uncertainty from the surrogate model.

\subsubsection{Acquisition Function}

After the surrogate model outputs predictive mean and uncertainty for
the unseen configurations, \SYSTEM{} needs an acquisition function to
select the best configuration to sample. A good acquisition function
should balance the tradeoffs between exploration and exploitation.
For \SYSTEM{}, we chose the expected improvement (EI) acquisition function, which has been demonstrated to
perform well in configuration
search~\cite{alipourfar2017cherrypick,patel2020clite,roy2021bliss}.
EI selects the configuration that would
  have the highest expected improvement over the best observed
  system behavior so far. Assuming $m_i$ is the best observed
  system behavior at the $i$-th round, we have
\begin{align}
\x_i = \argmax_{\x\in \mathbb{R}^p} \begin{cases}
\left(\mu_i(\x)-m_i \right)\Phi(Z)+\sigma_i(\x)\phi(Z), & \text{if } \sigma_i(\x)>0 \nonumber \\
0, & \text{if } \sigma_i(\x)=0
\end{cases}
\end{align}
where $Z=\frac{\mu_i(\x)-m_i}{\sigma_i(\x)}$, $\Phi(\cdot)$ is the
cumulative distribution function and $\phi(\cdot)$ is the probability
density function, both assuming normal distributions. In particular,
$\left(\mu_i(\x)-m_i \right)\Phi(Z)$ indicates the improvement in
favor of exploitation, and $\sigma_i(\x)\phi(Z)$ indicates the
uncertainty in favor of exploration. The unseen configuration with the
highest expected improvement will be selected.

\subsection{System Monitor}

When a configuration is selected, \SYSTEM{} runs the workload in this
configuration and monitors its status. The System Monitor (\circled{B}) records
behavior (i.e., latency and energy) at regular time intervals to check
if the configuration has finished running (\circled{1}). If the
configuration finishes, \SYSTEM{} reports the latency and energy back
to Bayesian Optimization (\circled{A}) for the next round.  If the
configuration is still running, the System Monitor reports the current
latency and energy to the next module.

\subsection{Censored Regressor}\label{sec:censor}

\SYSTEM{}'s Censored Regressor (\circled{C}) takes the information
from the System Monitor (\circled{B}) and predicts the current
configuration's final latency and energy.
The key idea of censored regression is to use the system
behavior at the current time interval to accurately predict whether
the current configuration will exceed the termination threshold. In
particular, censored regression categorizes the training samples as
follows:
\begin{itemize}
\item \rectangled{1} Samples that finish running; i.e.,
  those which have terminated normally and thus have known latency and
  energy. 
\item \rectangled{2} Samples that do not finish running; i.e., those which are terminated early and their latency and energy are censored at the current elapsed latency and energy.
\end{itemize}
Next, we describe how censored regression models these two
categories of training data. We will use the problem of minimizing
latency under a power constraint as an example, in which we will
predict latency.

Let $z_i$ be the observed latency associated with $\x_i$ at the $i$-th
round, and $g(\cdot)$ be the output from censored regression. Then the formal
  definition of censored regression model~\cite{barnwal2020survival} is:
\begin{align}\label{eq:z}
\log z_i = g(\x_{i}) + \varsigma\epsilon_i,
\end{align}
where $\varsigma$ is a constant and $\epsilon_i$ is the
measurement noise. The $\log$-scale of the response
  simplifies modeling different types of exponential distributions. The goal of censored regression is to estimate $g(\cdot)$. When $\x_i$ is running and the System Monitor returns the
current latency at the $t$-th time interval, the current latency
$y_{it}$ is a censored value of $z_i$ such that
\begin{align}\label{eq:current}
y_{it} = \begin{cases}
z_i,  &  z_i < \tau_{it} \quad (\rectangled{1}) \\
\tau_{it},  &   z_i \geq \tau_{it} \quad (\rectangled{2})
\end{cases}
\end{align}
where $\tau_{it}$ is the current latency at the $t$-th time
interval (i.e., the amount of time that has elapsed during the execution of the sample). In~\cref{eq:current}:
\begin{itemize}
\item If the observed latency is less than the current latency $z_i <
  \tau_{it}$, \SYSTEM{} observes the latency when the configuration finishes running; i.e., $y_{i\cdot} =
  z_i$, where $y_{i\cdot}$ is the observed latency of a sample that
  terminated normally. \SYSTEM{} treats these uncensored (finished) samples by fitting them with regression, corresponding to \rectangled{1}.
\item If the observed latency is greater than or equal to the current
  latency $z_i \geq \tau_{it}$, \SYSTEM{} observes the current
  latency censored at the $t$-th time interval, and $\tau_{it}$ is the
  current elapsed latency. \SYSTEM{} treats this censored (running) sample by estimating its probability of
  $z_i$ exceeding $\tau_{it}$ given the configuration, corresponding to \rectangled{2}.
\end{itemize}
Censored regression treats the uncensored (finished) samples and the censored (running) sample differently, and solves them together by minimizing the total negative log-likelihood (NLL) loss function. The NLL loss function at the $t$-th time interval in the $i$-th round is:
\begin{align}\label{eq:mle}
\rho\left(g;(\x_{j}, y_{j\cdot})_{j=1}^{i-1},(\x_{i},
\tau_{it})\right) &=-\sum_{j=1}^{i-1}\mathbbm{1}_{[z_{j}=y_{j\cdot}]}\log\frac{1}{y_{j\cdot}\sigma}\phi\left(\frac{\log y_{j\cdot}-g(\x_{j})}{\sigma}\right) \nonumber \\
&
-\log\left(1-\Phi\left(\frac{\log\tau_{it}-g(\x_{i})}{\sigma}\right)\right),
\end{align}
where $\phi(\cdot)$ and $\Phi(\cdot)$ are the probability density and
cumulative distribution functions of the noise $\epsilon_i$ in~\cref{eq:z}, respectively. The first term is the log-likelihood function for the uncensored (finished) samples that measures the
discrepancy between $\log y_{j\cdot}$ and $g(\x_{j})$ where $j\in[1,i-1]$, corresponding to
\rectangled{1}. The second term is the log-likelihood function for the
censored (running) sample of the current $i$-th round that
encourages $g(\x_{i})>\log\tau_{it}$ when $y_{it}=\tau_{it}$,
corresponding to \rectangled{2}. To fit this loss function, \SYSTEM{}
uses gradient boosting trees due to its high fitting power in various
predictive tasks~\cite{chen2016xgboost}. Because, all else equal, the loss function in \cref{eq:mle} is lower when $g(\x_{j})$ is higher, we regularize the predictor by training with early stopping~\cite[Chapter 5.5.2]{bishop2006pattern}.

\subsection{Predictive Early Termination}

\SYSTEM{}'s Predictive Early Termination Module (\circled{D}) will
terminate the current configuration based on the predictions
from the Censored Regressor (\circled{C}).  If the prediction
is better than the existing best from the collected samples, \SYSTEM{}
lets the configuration keep running until the next time interval, when
\SYSTEM{} will make another prediction to check if the configuration
should be terminated. If the prediction is worse than the
existing best from the collected samples, \SYSTEM{} will terminate the
running process, drop this configuration, and begin the next round of
Bayesian Optimization (\circled{A}). In this case, the surrogate model
does not change since no new samples are added to the training set. To
avoid selecting the same configuration as the previous rounds,
\SYSTEM{} only selects the configuration that maximizes the acquisition
function from the unsampled configurations. With this module, \SYSTEM{} reduces the
time cost of each sample by dynamically determining whether predictive early
termination should happen and when it will happen.

\begin{algorithm}[!htb]
	\begin{algorithmic}[1]
		\small
		\Require $T_{\rm budget}$ \Comment{Time budget.}
		\Require $C$ \Comment{Target constraint in power or latency.}
		\State Randomly sample a configuration $\x_0$ to construct $X_{\mathrm{train}}$ and obtain its  system behavior $y_0$ to construct $Y_{\mathrm{train}}$.
		\State $i=1$ 
		\While{$T_{\rm budget}>0$}
		\State Train surrogate model using $X_{\mathrm{train}}$ and $Y_{\mathrm{train}}$. \label{line:surrogate}
		\State Select configuration $\x_i$ based on acquisition function. \label{line:acquisition}
		\State Run experiment at $\x_i$ and monitor its system behavior. \label{line:run}
		\State $\gamma_i \gets \min Y_{\mathrm{train}}$.  \Comment{Update the existing best.} \label{line:update}
		\For{each interval $t=1,\ldots,$}
		\If {\rm sample not finished} \label{line:not-c}
		\State Get current system behavior $y_{it}$. \label{line:censor-1}
		\State Build censored regressor on $X_{\mathrm{train}}$, $Y_{\mathrm{train}}$, $\x_i$, $y_{it}$.\label{line:censor-2}
		\State Get predicted latency $\hat{y}_{it}$ for $\x_{i}$. \label{line:censor-3}
		\If {$\hat{y}_{it} \geq \gamma_i$} \label{line:terminate-1} 
		\State Early terminate experiment.\label{line:terminate-2}
		\State Get current running time $t_{it}$.
		\State $T_{\rm budget} \gets T_{\rm budget} - t_{it}$. \label{line:time-1}
		\Else
		\State Get system behavior $y_{i\cdot}$ and constraint $c_i$ for $\x_i$. \label{line:con-1}
		\If{$c_i \leq C$ } \label{line:con-2}
		\State $X_{\mathrm{train}}\gets X_{\mathrm{train}} \bigcup \x_i$, $Y_{\mathrm{train}}\gets Y_{\mathrm{train}}\bigcup y_{i\cdot}$. \label{line:con-3}
		\Else  
		\State $X_{\mathrm{train}}\gets X_{\mathrm{train}} \bigcup \x_i$, $Y_{\mathrm{train}}\gets Y_{\mathrm{train}}\bigcup \mathrm{inf}$. \label{line:con-4}
		\EndIf 
		\State Get current running time $t_{it}$.
		\State $T_{\rm budget} \gets T_{\rm budget} - t_{it}$. \label{line:time-2}
		\EndIf
		\EndIf
		\EndFor
		\State $i \gets i+1$.
		\EndWhile
		\State \Return $\x^\star$ with the best system behavior from $X_{\mathrm{train}}$ that meets the target constraint.  \label{line:out}
	\end{algorithmic}
	\caption{\SYSTEM{} for systems optimization.}
	\label{alg:cello}
\end{algorithm}

\subsection{Applying \SYSTEM{} to Optimization Problems}

\SYSTEM{} is a general framework that can augment existing SEML approaches by reducing the time cost per sample. We apply \SYSTEM{} to solve the following two optimization problems:
(1) \textbf{\LUP{}:} minimizing latency under a power constraint; and (2)
\textbf{\EUL{}:} minimizing energy under a latency constraint.

Algorithm~\ref{alg:cello} summarizes the procedure. In
\cref{line:surrogate,line:acquisition}, \SYSTEM{} conducts Bayesian
Optimization to select the configuration to run (\circled{A}). In
\cref{line:run}, \SYSTEM{}'s System Monitor observes the running
configuration (\circled{B}). In \cref{line:update}, \SYSTEM{} updates
the existing best system behavior (latency for \LUP{}, energy for
\EUL{}) from the collected samples. At each time interval, if the
sample has not finished (\cref{line:not-c}), \SYSTEM{}'s Censored
Regressor (\circled{C}) predicts future behavior
(\cref{line:censor-1,line:censor-2,line:censor-3}). If the predicted
behavior is worse than the existing best (\cref{line:terminate-1}),
\SYSTEM{} terminates the configuration (\circled{D},
\cref{line:terminate-2}), and updates the time budget
(\cref{line:time-1}). If the experiment has terminated normally,
\SYSTEM{} determines how to update the training set. If the
configuration meets the target constraint---in either latency or
power---(\cref{line:con-2}), \SYSTEM{} adds this configuration and its
true behavior to the training set for the next round
(\cref{line:con-3}). Otherwise, \SYSTEM{} adds this configuration and
uses a extremely high value to replace its true behavior
(\cref{line:con-4}), which incorporates the knowledge that this
configuration violates the target constraint by adding a discontinuous
sample to the training set. Meanwhile, \SYSTEM{} updates the time
budget (\cref{line:time-2}) to make sure that there is enough time to
go to the next round. Finally, \SYSTEM{} outputs the best
configuration that meets the target constraint (\cref{line:out}). We
implement \SYSTEM{} in Python with libraries including
numpy~\cite{numpy}, pandas~\cite{pandas}, and
scikit-learn~\cite{scikit-learn}. For censored regression, we use an
implementation from the Accelerated Failure Time model in XGBoost
library, which is a state-of-the-art implementation of censored
regression fitted with the gradient boosting
trees~\cite{barnwal2020survival}. The code is released
in~\url{https://anonymous.4open.science/r/cello-code-8512}.

\begin{table*}[!htb]
	\small
	\begin{center}
		\caption{Hardware and software configuration parameters tuned in the experiments.}
		\begin{tabular}{clll}
			\toprule
			\textbf{Category} & \textbf{Configuration parameter}        & \textbf{Range} & \textbf{Description}                                                          \\ \midrule	
			\multirow{5}{*}{\rotatebox{90}{Hardware}} & cpu.freq                                & 1.0--3.7       & CPU frequency, in GHz.                                                        \\
			& uncore.freq                             & 1.0--2.4       & Uncore frequency, in GHz.                                                     \\
			& hyperthreading                          & on, off        & Hyperthreading.                                                               \\
			& n.sockets                               & 1, 2           & Number of sockets.                                                            \\
			& n.cores                            & 1--12          & Number of cores per socket.                                                   \\\midrule	
			\multirow{20}{*}{\rotatebox{90}{Software}}   & spark.reducer.maxSizeInFlight           & 24--128        & Max size of map outputs to fetch from each reduce task, in MB. \\
			& spark.shuffle.file.buffer               & 24--128        & Size of the in-memory buffer for each shuffle file output stream, in KB.      \\
			& spark.shuffle.sort.bypassMergeThreshold & 100--1000      & Avoid merge-sorting data if there is no map-side aggregation.                 \\
			& spark.speculation.interval              & 100--1000      & How often Spark will check for tasks to speculate, in millisecond.            \\
			& spark.speculation.multiplier            & 1--5           & How many times slower a task is than median considered for speculation. \\
			& spark.speculation.quantile              & 0--1           & Percentage of tasks to be complete before speculation is enabled.     \\
			& spark.broadcast.blockSize               & 2--128         & Size of each piece of a block for TorrentBroadcastFactory, in MB.             \\
			& spark.io.compression.snappy.blockSize   & 24--128        & Block size used in snappy, in KB.                                             \\
			& spark.kryoserializer.buffer.max         & 24-128         & Maximum allowable size of Kryo serialization buffer, in MB.                   \\
			& spark.kryoserializer.buffer             & 24--128        & Initial size of Kryo’s serialization buffer, in KB.                           \\
			& spark.driver.memory                     & 6--12          & Amount of memory to use for the driver process, in GB.                        \\
			& spark.executor.memory                   & 6--16          & Amount of memory to use per executor process, in GB.                          \\
			& spark.network.timeout                   & 20--500        & Default timeout for all network interactions, in second.                      \\
			& spark.locality.wait                     & 1--10          & How long to launch a data-local task before giving up, in second.             \\
			& spark.task.maxFailures                  & 1--8           & Number of task failures before giving up on the job.                          \\
			& spark.shuffle.compress                  & false, true    & Whether to compress map output files.                                         \\
			& spark.memory.fraction                   & 0--1           & Fraction of (heap space--300 MB) used for execution and storage.              \\
			& spark.shuffle.spill.compress            & false, true    & Whether to compress data spilled during shuffles.                             \\
			& spark.broadcast.compress                & false, true    & Whether to compress broadcast variables before sending them.                  \\
			& spark.memory.storageFraction            & 0.5--1         & Amount of storage memory immune to eviction.                                  \\ \bottomrule
		\end{tabular}\label{tbl:conf}
	\end{center} 
\end{table*}

%% file: tex/evaluation.tex
\section{Experimental Setup} \label{sec:setup}

\subsection{Software Systems}\label{sec:sw-systems}

We use Apache Spark as our software system~\cite{spark}. Each
experiment has a master node and four worker nodes. We use a wide
range of configuration parameters that reflect significant properties
categorized by shuffle behavior, data compression and serialization,
memory management, execution behavior, networking, and scheduling
(details in the software category of Table~\ref{tbl:conf}). We use the
same set of parameters as prior work~\cite{yu2018datasize} slightly
modified for the current version of HiBench Spark.

We use 12 workloads (Table~\ref{tbl:workloads}) from
HiBench's big data benchmark suite~\cite{huang2010hibench}---a
benchmark suite that is widely used in prior
work~\cite{yu2018datasize,wang2019tpshare,banerjee2021bayesperf,ding2021generalizable}.
The workloads cover various domains including microbenchmarks, machine
learning, websearch, and graph analytics.

\begin{table}[!htb]
	\begin{center}
		\caption{HiBench workloads.} 
		\begin{tabular}{ll|ll} \toprule
			\textbf{Workload}   & \textbf{Data size} & \textbf{Workload} &  \textbf{Data size}  \\
			\midrule
			als         & 0.6 GB  & bayes       & 19 GB    \\
			gbt         & 2 GB    & kmeans      & 20 GB    \\
			linear      & 48 GB   & lr          & 8 GB     \\
			nweight     & 0.9 GB  & pagerank    & 1.5 GB   \\
			pca         & 4 GB    & rf          & 0.8 GB   \\
			terasort    & 3.2 GB  & wordcount   & 32 GB    \\
			\bottomrule   
		\end{tabular}
		\label{tbl:workloads}
	\end{center} 
\end{table}

\subsection{Hardware Systems}\label{sec:hw-systems}

We run experiments on the Chameleon configurable cloud computing
systems~\cite{keahey2020lessons}, where each node is a dual-socket
system running Ubuntu 18.04 (GNU/Linux 5.4) with 2 Intel(R) Xeon(R)
Gold 6126 processors, 192 GB of RAM, hyperthreads and TurboBoost. Each
socket has 12 cores/24 hyperthreads and a 20 MB last-level cache. The
hardware parameters we optimize are in the hardware
category of Table~\ref{tbl:conf}, which have been shown to have a
significant affect on latency and power
tradeoffs~\cite{zhang2016maximizing}.

\subsection{Points of Comparison}\label{sec:comparison}

We compare a spectrum of approaches including random sampling
(\RS{}), Bayesian optimization (\BO{}), Bayesian optimization with
measured early termination (\BOST{}, \BOTC{}, \BOIM{}, \BONN{}) and
predictive early termination (\BOGB{}), and ensemble Bayesian
optimization (\Bliss{}).
\begin{itemize}
\item \textbf{\RS{}}: randomly sample configurations and select the
  best configuration that meets the constraint.
\item \textbf{\BO{}}: Bayesian optimization using random forest as the
  surrogate model and expected improvement as the acquisition
  function~\cite{alipourfar2017cherrypick,nardi2019practical,patel2020clite,chen2021efficient}.
  We also use the same setting for the following approaches.
\item \textbf{\BOST{}}: Bayesian optimization with measured early
  termination using the static termination threshold; i.e., terminate
  the experiment when the measured latency or energy is observed to
  reach the static termination threshold, and then drop the sample.
  The static threshold is the latency or energy of the initial sample
  because it is the only prior knowledge we have to set this
  threshold.
\item \textbf{\BOTC{}}: Bayesian optimization with measured early
  termination by truncation; i.e., terminate the experiment when the
  measured latency or energy is observed to be worse than the existing
  best, and then drop the configuration.
\item \textbf{\BOIM{}}: Bayesian optimization with measured early
  termination by imputation~\cite{hutter2013bayesian}; i.e., terminate
  the experiment when the measured latency or energy is observed to be
  worse than the existing best, and then impute---i.e., replace
  missing data with predicted values---the unfinished configuration.
\item \textbf{\BONN{}}: Neural model-based Bayesian optimization with
  measured early termination and censored
  observations~\cite{eggensperger2020neural}; i.e., train the neural
  network with the Tobit loss function for the censored training data.
  The same neural network architecture and hyperparameter settings are
  used in~\cite{eggensperger2020neural}. We use the same static
  threshold setting as the experiments
  in~\cite{eggensperger2020neural}.  The threshold is the latency or
  energy of the initial sample because it is the only prior knowledge
  we have to set this threshold.
\item \textbf{\BOGB{}}: Bayesian optimization with predictive early
  termination by predicting with standard regression; i.e., terminate
  the experiment when the predicted latency or energy is worse than
  the existing best, and drop the configuration. We use gradient
  boosting trees as the standard regressor due to its high
  accuracy~\cite{chen2016xgboost}. This approach is novel to
  this paper, we include it to show the importance of using censored
  regression rather than standard regression for predicted early
  termination.
\item \textbf{\Bliss{}}: ensemble Bayesian
  optimization~\cite{roy2021bliss}.  \Bliss{} reduces sample
  collection time by determining when to update its surrogate model
  using predictions from that same surrogate.  To the best of our
  knowledge, \Bliss{} is the state-of-the-art of Bayesian optimization
  in configuration search for computer systems.
\item \textbf{\SYSTEM{}}: Bayesian optimization with predictive early
termination by predicting with censored regression.
\end{itemize}

\subsection{Evaluation Methodology}\label{sec:eval-method}

Following a methodology established in prior work on investigating
Apache Spark systems~\cite{yu2018datasize}, we create a set of 2000
configurations per workload, randomly sampled from both hardware and
software configuration parameters. We run these configurations to
record their latency and energy consumption. For reference, we used
more than 50000 CPU hours to record this test dataset (and we will
include it as part of our open source release). We then give each of
the above approaches a fixed amount of time to search a combined space
of hardware and software configuration parameters and find solutions
for two optimization problems: (1) \textbf{\LUP{}:} minimizing latency
under a power constraint; and (2) \textbf{\EUL{}:} minimizing energy
under a latency constraint.

We evaluate on a wide range of search time budgets such that at least
one of the above approaches converges to the optimal solution. We set
a range of constraints: the power and latency constraints are set as
$[10, 20, 30, 40, 50, 60, 70, 80, 90]$-th percentiles of the
distributions. For all approaches that measure progress of running
samples, we set a time interval of 5 seconds; i.e., the system
behavior is monitored every 5 seconds.  The reported results are
averaged over different constraints over 10 runs with different random
seeds.

For censored regression from the XGBoost library, we choose the
\texttt{extreme} distribution for the loss function. We set
\texttt{num\_boost\_round}, the parameter that controls the number of
steps to fit gradient boosting trees in~\cref{eq:mle} to minimize the
loss (i.e., the point at which to early stop) to 20. We tune this
parameter by generating predictions that are generally above the
current elapsed latency and energy on one workload \texttt{als}
dataset, and use this same value on all other datasets without
additional tuning. For each distinct trial on each dataset, we run
cross validation for \texttt{distribution\_scale} with the value range
of $[0.2, 0.3, 0.4]$ and \texttt{learning\_rate} with the value range
of $[0.2, 0.25, 0.3]$ and report the best results.

\subsection{Evaluation Metric}

\noindent \textbf{RE.} We use relative error (RE) between the result from
each approach and the optimal for evaluation:
\begin{align}
\text{RE} = \frac{|Y_{\rm pred} - Y_{\rm opt}|}{|Y_{\rm opt}|},
\label{eq:ape}
\end{align}
where $Y_{\rm pred}$ is the best value found by the approach, and
$Y_{\rm opt}$ is the optimal measured value. Lower RE is better.

\section{Experimental Evaluation}\label{sec:evaluation}

\begin{figure*}[!htb]
	\centering
	\includegraphics[width=\linewidth]{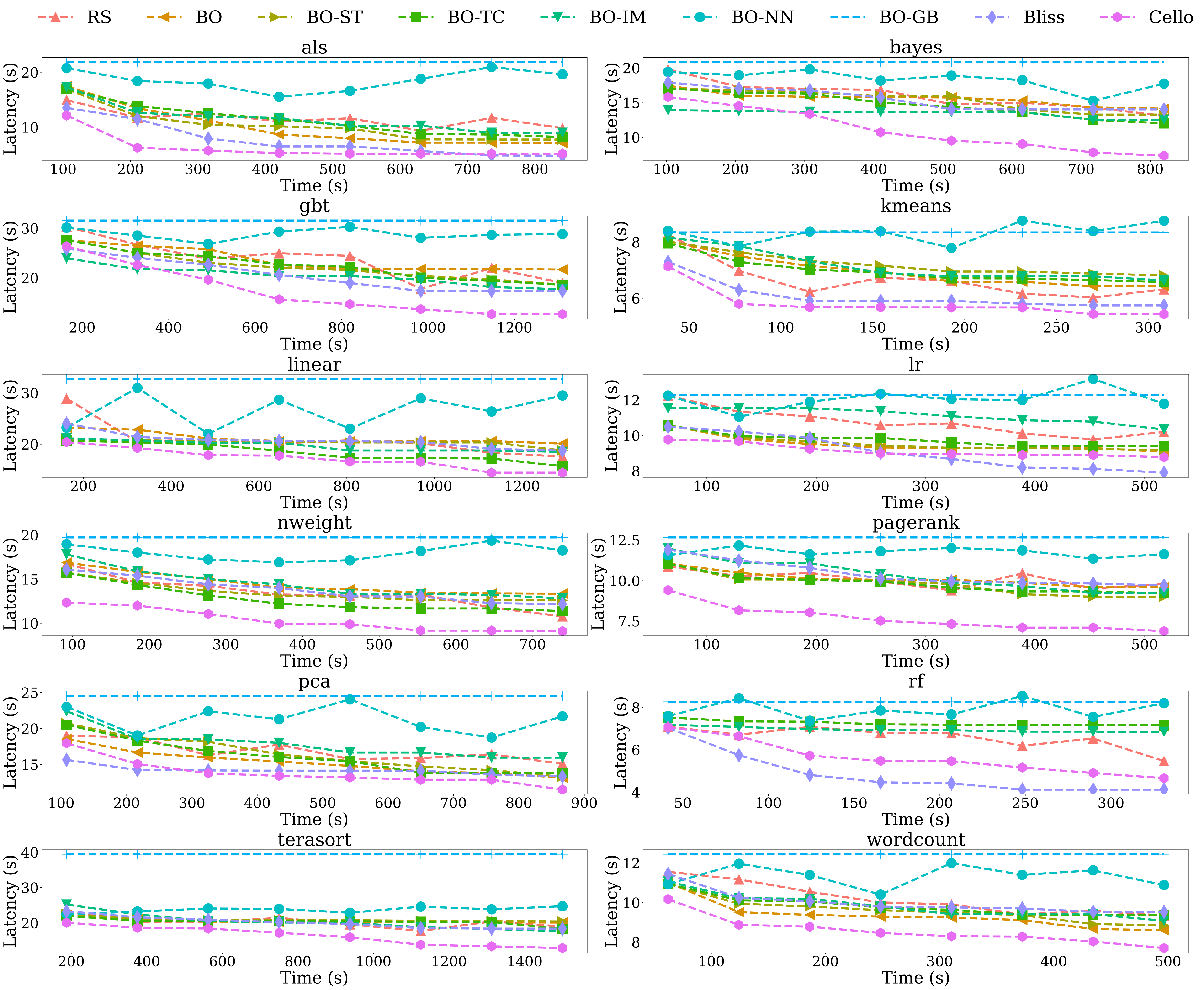} 
	\caption{Latency results averaged over all power constraints
          for \LUP{} (lower is better).}
	\label{fig:pup}
\end{figure*}

\begin{figure*}[!htb]
	\centering
	\includegraphics[width=\linewidth]{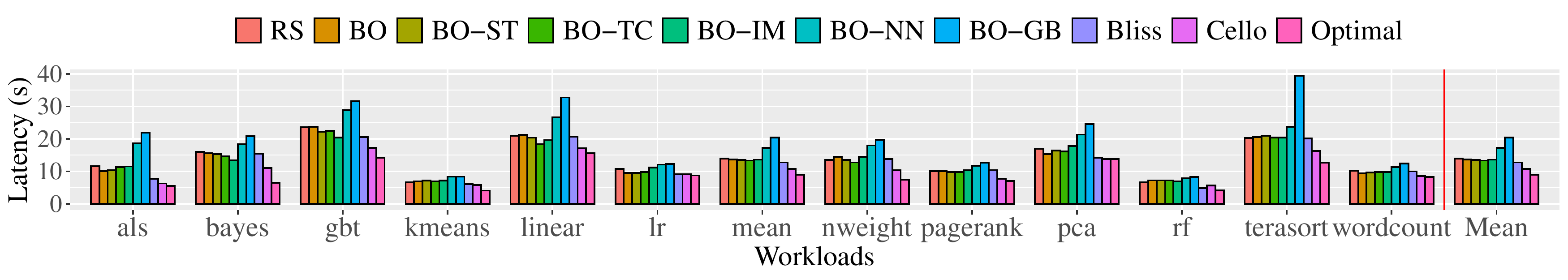}
	\caption{Latency results averaged over all search time budgets
          for \LUP{} (lower is better).}\label{fig:bar-pup}
\end{figure*}

\begin{figure*}[!htb]
	\centering
	\includegraphics[width=\linewidth]{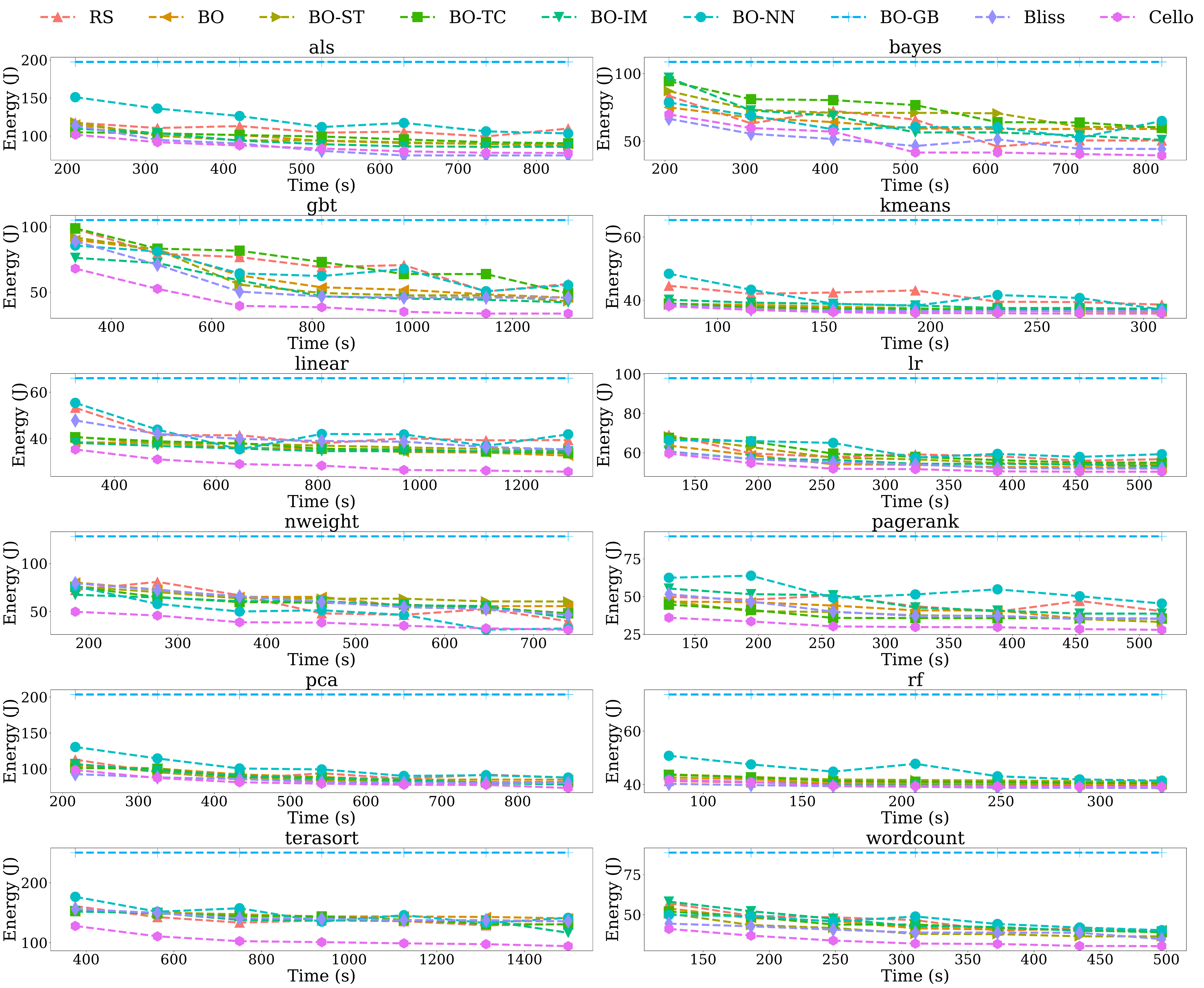} 
	\caption{Energy results averaged over all latency constraints for \EUL{}s (lower is better).}  
\label{fig:eup}
\end{figure*}

\begin{figure*}[!htb]
	\centering
	\includegraphics[width=\linewidth]{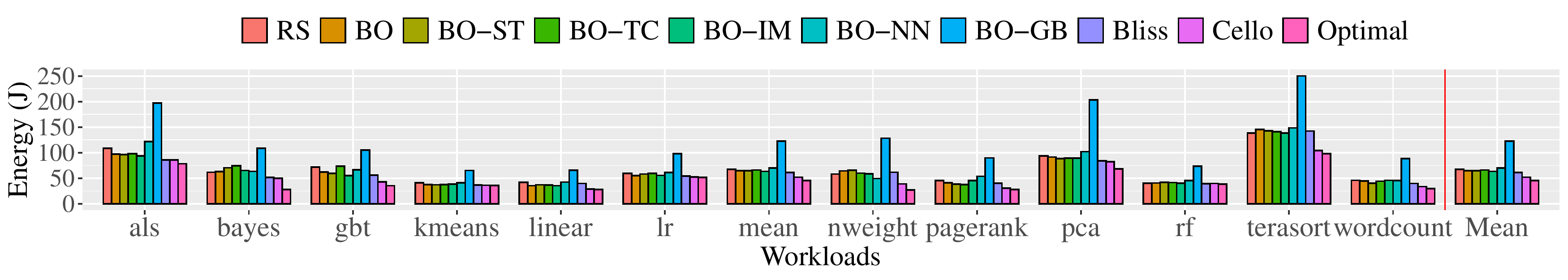} 
	\caption{Energy results averaged over all search time budgets for \EUL{} (lower is better).}\label{fig:bar-eup}
\end{figure*}

We evaluate the following research questions (RQs):
\begin{itemize}
	\item \textbf{RQ1:} How well does \SYSTEM{} perform?
	\SYSTEM{} reduces latency by 1.19--1.90$\times$ for \LUP{}
		(Figure~\ref{fig:pup} and~\ref{fig:bar-pup}) and energy by
		1.18--2.35$\times$ for \EUL{} (Figure~\ref{fig:eup}
		and~\ref{fig:bar-eup}).
	\item \textbf{RQ2:} Is prediction or measurement better for early termination?
	\SYSTEM{}'s predictive early termination outperforms measured
		early termination approaches by terminating much earlier while
		producing more accurate predictions (\cref{sec:res-pred}).
	\item \textbf{RQ3:} How is censored regression beneficial to \SYSTEM{}?
 \SYSTEM{}'s censored regression reduces prediction error (as
		mean squared error) for latency and energy by 60\% and 82\% over
		standard regression for \LUP{} and \EUL{} (Figure~\ref{fig:gbcr}).
	\item \textbf{RQ4:} How many samples are explored?
By combining censored regression and predictive early
		termination, \SYSTEM{} both explores more samples and improves search
		results compared to other baselines (\cref{sec:res-samples}).
	\item \textbf{RQ5:} What is the overhead?
Within the same amount of search time (learning overhead
		included), \SYSTEM{} produces the best results by both predicting
		accurately (\cref{sec:res-censor}) and updating quickly (0.3 seconds
		per sample, \cref{sec:res-overhead}).
\end{itemize}

\subsection{RQ1: How well does \SYSTEM{} perform?}\label{sec:res-good}

Figure~\ref{fig:pup} and~\ref{fig:eup} show the latency and energy
results of different approaches as a function of different search time
budgets for \LUP{} and \EUL{}, respectively, where the x-axis is the
search time, and the y-axis is the latency or energy (lower is
better). We compare the summarized results averaged over all search
times in Figure~\ref{fig:bar-pup} and~\ref{fig:bar-eup}, and compute
the relative error (RE) by comparing to the optimal value (Optimal) in
Table~\ref{tbl:main-res}, where Optimal is the best value found
through brute force search. We can see that compared to other
approaches, \SYSTEM{} finds the configuration with the lowest latency
and energy at almost all search times for almost all workloads. The
major exceptions are \texttt{lr} and \texttt{rf} in \LUP{}, and
\texttt{als} and \texttt{bayes} in \EUL{}, where \Bliss{} outperforms
\SYSTEM{} at some points. It is not surprising since with numerous
local optima in the tremendous search space, no single method will
dominate. Despite that, \SYSTEM{} still outperforms prior work in the
majority of cases and gets much better results on average. In
particular, we find that on average:

\begin{table}[!htb] 
	\small
	\begin{center}
		\caption{Summarized results averaged over all workloads for each approach. Lower RE is better.}
		\begin{tabular}{l|ll|ll} \toprule
			& \multicolumn{2}{c}{\LUP{}} & \multicolumn{2}{c}{\EUL{}} \\ \midrule
			& RE (\%)        & \# samples         & RE (\%)       & \# samples         \\ \midrule
			RS    & 55  & 23  & 46 & 25  \\
			BO    & 52  & 24  & 36 & 28  \\
			BO-ST & 51  & 26  & 31 & 32  \\
			BO-TC & 48  & 30  & 32 & 41  \\
			BO-IM & 51  & 91  & 31 & 69  \\
			BO-NN & 94  & 19  & 54 & 21  \\
			BO-GB & 127 & 385 & 72 & 378 \\
			Bliss & 42  & 36  & 31 & 31  \\
			Cello & 20  & 44  & 16 & 50 \\ \bottomrule    
		\end{tabular} \label{tbl:main-res}
	\end{center}
\end{table}

\begin{itemize}
\item \SYSTEM{} achieves 20\% and 16\% REs for \LUP{} and \EUL{}
  respectively (Table~\ref{tbl:main-res}), which represent
  1.19$\times$ and 1.18$\times$ speedups compared to the best baseline
  \Bliss{}, and 1.90$\times$ and 2.35$\times$ speedups compared to the
  weakest baseline \BOGB{}.
\item \BO{} underperforms all approaches with early termination except
  \BOGB{}, which indicates the benefits of early terminations in SEML.
  The reasons that \BOGB{} performs poorly will be discussed in
  \cref{sec:res-censor}.
\item As the state-of-the-art Bayesian optimization approach for
  configuration optimization, \Bliss{} under-performs \SYSTEM{} in
  most workloads for several reasons. First, \Bliss{} does not
  incorporate early termination; it runs a few rounds as \BO{} in the
  beginning and then starts using predictions to update the surrogate
  model, which means that it does not reduce the cost of collecting
  samples during early rounds. Second, \Bliss{} uses its surrogate
  model---i.e., standard regression---to predict, which leads to less
  accurate prediction compared to censored regression (more details
  in~\cref{sec:res-censor}). Finally, \Bliss{} updates its surrogate
  alternating between using predictions and measured system behavior,
  which produces less accurate model updates.  In contrast, \SYSTEM{}
  combines censored regression and early termination based on
  predicted behavior to reduce search costs.  When running for a fixed
  time budget, \SYSTEM{} explores more configurations than \Bliss{}
  (\texttt{\# sample} columns in Table~\ref{tbl:main-res}), and thus
  generally achieves better results.
\end{itemize}

\subsection{RQ2: Is prediction or measurement better for early
  termination?}\label{sec:res-pred}

Here, we focus on comparing predictive early termination (\SYSTEM{})
to measured early termination (\BOST{}, \BOTC{}, \BOIM{} and \BONN{})
to determine if the former significantly improves early termination
methods.  In Figure~\ref{fig:bar-pup} and~\ref{fig:bar-eup} ,
\SYSTEM{} has 1.24--1.61$\times$ and 1.24--1.34$\times$ speedups over
\BOST{}, \BOTC{}, \BOIM{} and \BONN{} for \LUP{} and \EUL{}
respectively.  \SYSTEM{} outperforms other measured early termination
approaches because it terminates poor samples much earlier using the
accurate predictions from censored regression. Therefore, given the
same amount of the search time budget, \SYSTEM{} also evaluate more
samples than \BOST{}, \BOTC{}, and \BONN{} from
Table~\ref{tbl:main-res}, which reflects the fact that \SYSTEM{}
terminates more samples earlier.  Although \BOIM{} evaluates more
samples than \SYSTEM{}, it underperforms \SYSTEM{}, which suggests
that it is terminating both early and accurately that improve results,
not just terminating early.  These results demonstrate the benefits of
predictive early termination over measured early termination; however,
the next section shows that achieving these benefits requires an
appropriate predictive model.

\begin{figure*}[!t]
	\centering
	\begin{subfloat}{\label{fig:gbcr-pup}
			\includegraphics[width=0.49\linewidth]{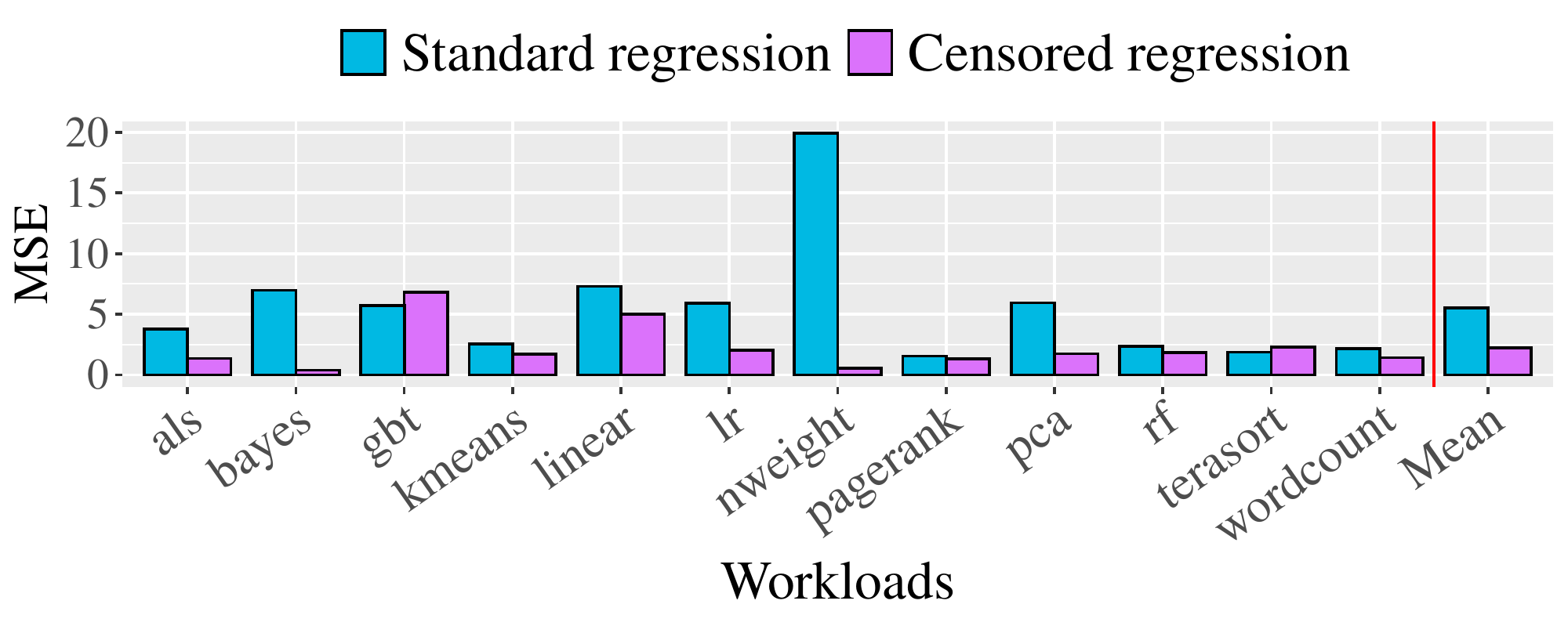}}	
	\end{subfloat}
	\begin{subfloat}{\label{fig:gbcr-eup}
			\includegraphics[width=0.49\linewidth]{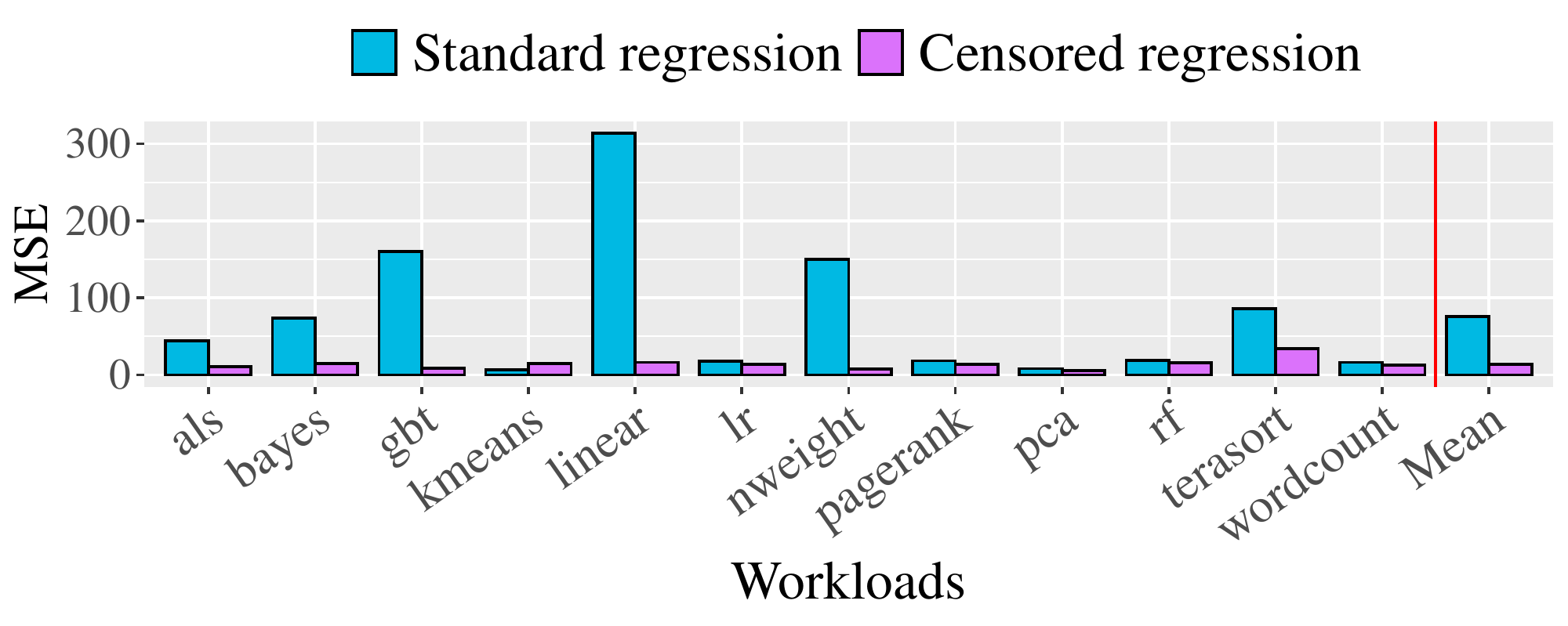}}		
	\end{subfloat}
	\caption{MSEs for \LUP{} (left) and \EUL{} (right). Lower is better.} 
	\label{fig:gbcr}
\end{figure*}

\begin{figure*}[!htb]
	\centering
	\includegraphics[width=\linewidth]{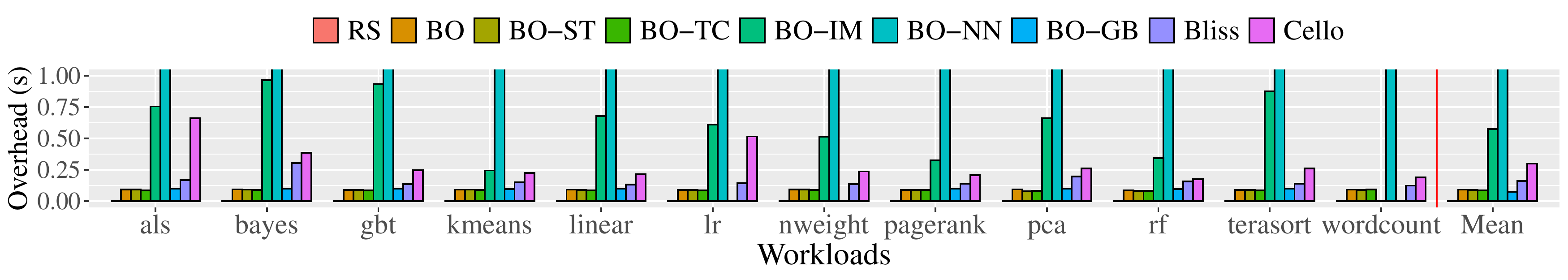} 
	\caption{Overhead of processing each sample for \LUP{}. Lower is better.}\label{fig:oh-pup}
\end{figure*}

\begin{figure*}[!htb]
	\centering
	\includegraphics[width=\linewidth]{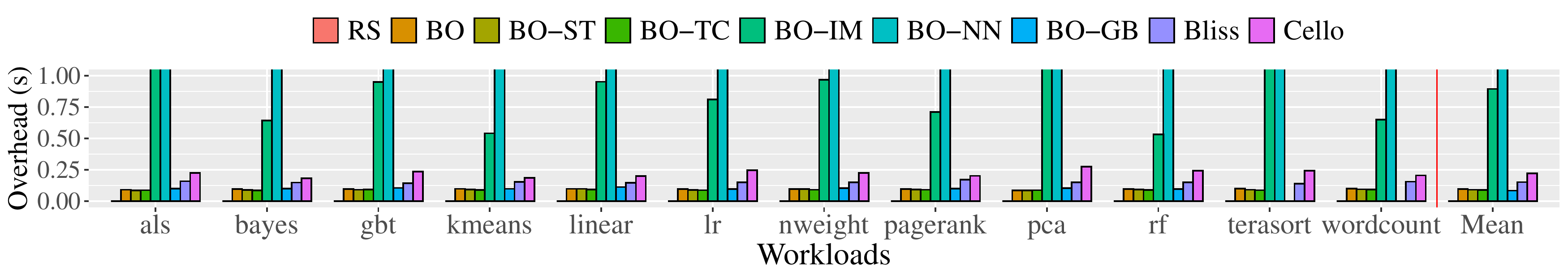} 
	\caption{Overhead of processing each sample for \EUL{}. Lower is better.}\label{fig:oh-eup}
\end{figure*}

\subsection{RQ3: How is censored regression beneficial to
  \SYSTEM{}?}\label{sec:res-censor}

Here, we analyze the results of \BOGB{} to \SYSTEM{}.
\BOGB{} uses early termination, but based on a standard regression
model rather than \SYSTEM{}'s censored regression.  Although \BOGB{}
has predictive early termination, it achieves the worst results.  This
poor result is due to the low prediction accuracy from standard
regression when training on censored data. To illustrate this, we
compare the prediction accuracy of each selected sample between
\BOGB{}'s standard regression (gradient boosting trees) and
\SYSTEM{}'s censored regression.

We monitor workload execution and predict its latency at each time
interval (\cref{sec:eval-method}). The censoring threshold for
censored regression is thus the elapsed latency observed at each time
interval. We compute the mean squared error (MSE) between predicted
latency and true latency.  Figure~\ref{fig:gbcr} shows the average MSE
over all time intervals, where the x-axis is the workload, the y-axis
is MSE (lower is better), and the last column, Mean, is the arithmetic
mean over all workloads.  Censored regression almost always has much
lower MSE than standard regression, with 60\% and 82\% reductions for
\LUP{} and \EUL{} respectively. Standard regression has worse
prediction accuracy because \BOGB{} terminates all samples throughout
the execution and thus does not update the surrogate model at all.
These results suggest that it is important to make accurate
predictions so that the poor configurations can be quickly identified.
Censored regression achieves high accuracy by considering both
uncensored (finished) and censored (running) samples, while standard
regression only takes uncensored (finished) samples into account.

\subsection{RQ4: How many samples are
  explored?}\label{sec:res-samples}

Table~\ref{tbl:main-res} summarizes the average number of samples
selected for each approach. Note that since all approaches are given
the same time to explore, exploring more samples is generally better,
provided those samples produce useful information for updating the
model.  We find that:
\begin{itemize}
\item Within the same time budgets, \BO{} selects fewer samples than
  others (except \BONN{}) as \BO{} does not use early termination to
  save extra time for more samples.
\item \BONN{} selects the fewest samples compared to others; 19 and 21
  on average for for \LUP{} and \EUL{} respectively. The reason is
  that \BONN{} has a surprisingly high overhead for model
  updates--over 11 seconds per sample (\cref{sec:res-overhead})---and
  thus a significant portion of its time is spent on surrogate updates
  instead of searching new configurations.
\item \BOGB{} selects the most samples; in Table~\ref{tbl:main-res},
  \BOGB{} selects 385 and 378 samples for \LUP{} and \EUL{}
  respectively. It is because \BOGB{} terminates all samples in the
  first time interval based on its predictions, and does not update
  the surrogate model at all.
  More results can be found in~\cref{sec:res-censor}.
\item \SYSTEM{} selects 44 and 50 samples for \LUP{} and \EUL{}
  respectively, which are more than \RS{}, \BO{}, \BOST{}, \BOTC{},
  \BONN{}, and \Bliss{}, while fewer than \BOIM{} and \BOGB{}. It
  indicates that increasing the number of samples is not the key to
  achieving the best results, but selecting more useful samples
  matters. \SYSTEM{} is such an example that combines early
  termination and censored regression to search more high-quality
  samples.
\end{itemize}

\subsection{RQ5: What is the overhead?}\label{sec:res-overhead}

We report the learning overhead of processing each sample, including
predicting future behavior and updating the surrogate.
Figure~\ref{fig:oh-pup} and~\ref{fig:oh-eup} show the overhead of
processing each sample for different approaches for \LUP{} and \EUL{}
respectively, where the x-axis is the workload, the y-axis is the
overhead, and the last column Mean is the arithmetic mean over all
workloads. We use the same setting for neural network training as
in~\cite{eggensperger2020neural} for \BONN{}, and \BONN{} has the
highest overhead; i.e., over 11 seconds on average. To better
visualize other methods properly, we cap the y-axis at 1 second.
\BOIM{} has the second largest overhead due to the high overhead of
imputing samples using the Expected-maximization
algorithm~\cite{hutter2013bayesian}.  The Expected-maximization
algorithm is an iterative method and thus needs numerous steps to
converge~\cite{moon1996expectation}.  \SYSTEM{} has the third largest
overhead because it uses the gradient boosting trees---which is an
ensemble method---to fit the loss function of censored regression.
Nevertheless, \SYSTEM{}'s overhead (0.3 seconds on average) is
negligible compared to the cost of collecting each sample (22.3
seconds on average), which is key for its applicability to the
efficient computer systems optimization problems.  All results
in this section include learning overhead for all approaches, and
\SYSTEM{} still produces the best results, showing that the overhead
can be negligible compared to making good decisions about early
termination to explore more high-quality samples in the same amount of
time.

\section{Discussion}
  
\begin{figure}[!htb]
	\centering
	\includegraphics[width=\linewidth]{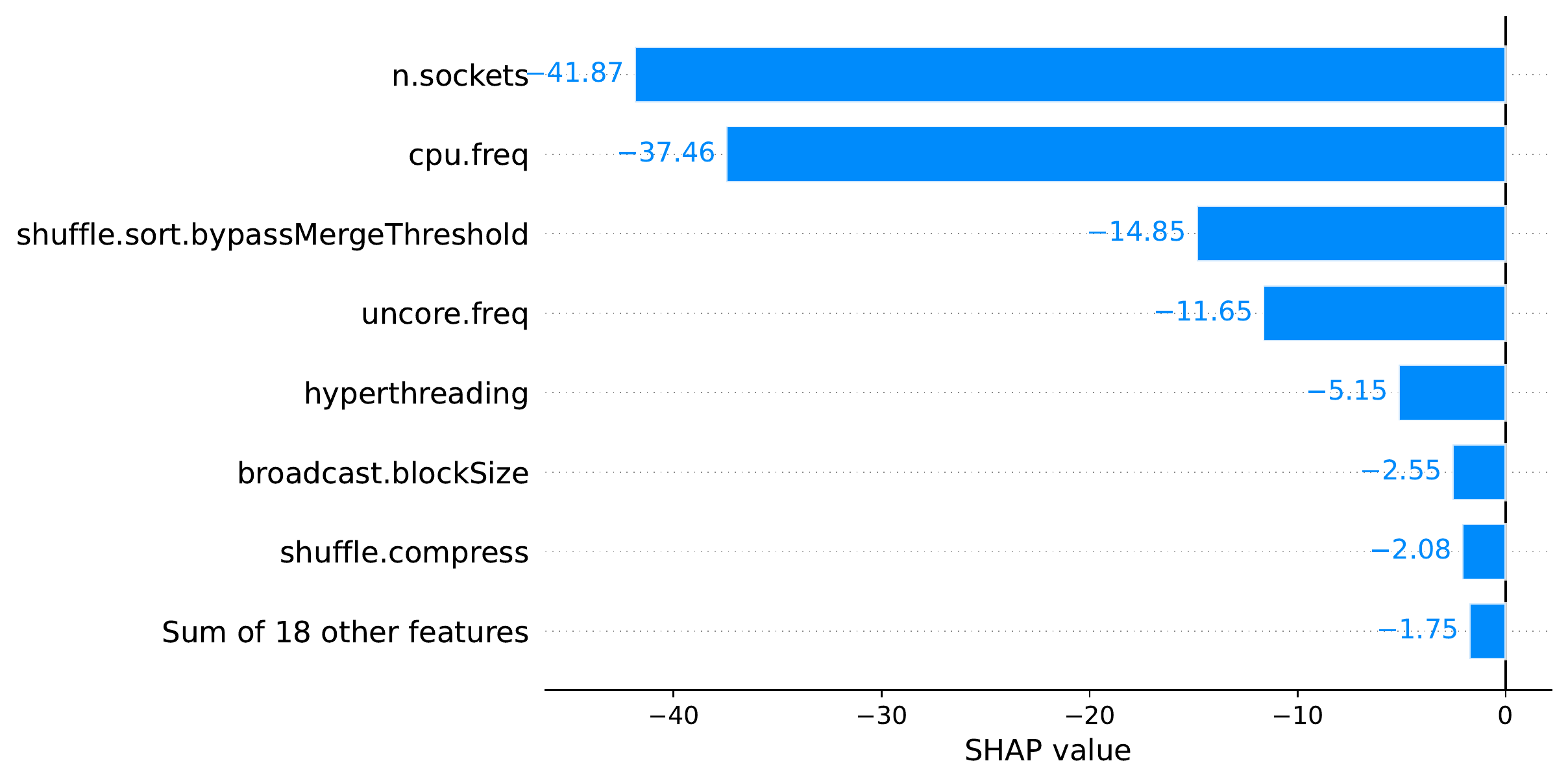} 
	\caption{SHAP values for the \texttt{als} workload. }\label{fig:shap_eup_als}
\end{figure}

\begin{figure}[!htb]
	\centering
	\includegraphics[width=\linewidth]{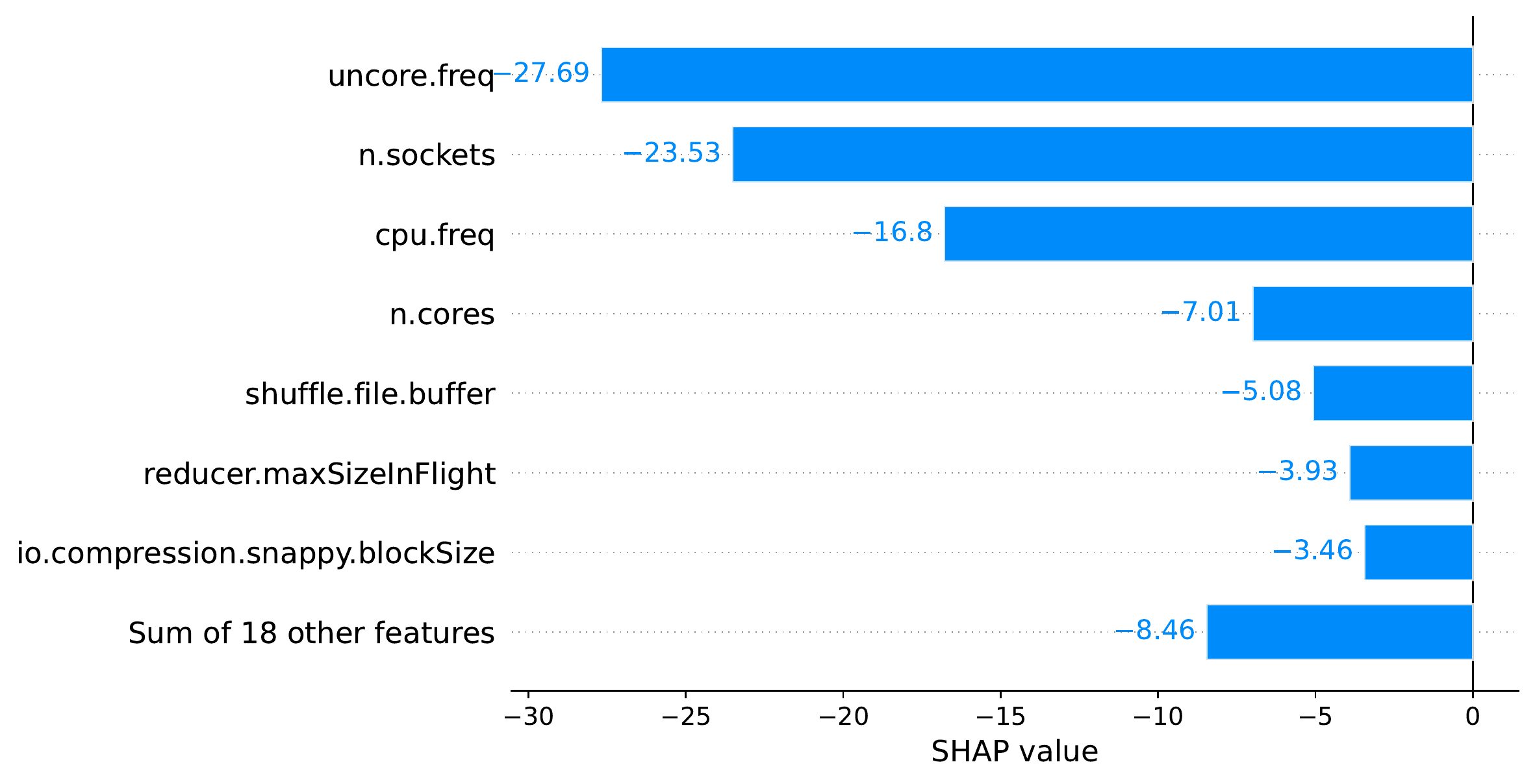} \vspace{-0.25in}
	\caption{SHAP values for the \texttt{bayes} workload. }\label{fig:shap_eup_bayes}
\end{figure}

To understand the impact of each configuration parameter on system
design, we visualize the SHAP values of the parameters with the
best energy under 50-th latency constraint for \EUL{} selected by
\SYSTEM{} on \texttt{als} and \texttt{bayes} workloads
in Figure~\ref{fig:shap_eup_als} and~\ref{fig:shap_eup_bayes}, respectively. SHAP
(SHapley Additive exPlanations) is a state-of-the-art approach to
interpret the output of machine learning models~\cite{lundberg2017unified}, which visualizes parameters contributing to push the model output from the base value (i.e., the average
model output over the training data we passed) to the model output.

For \texttt{als} in Figure~\ref{fig:shap_eup_als}, \texttt{n.socket} and \texttt{cpu.freq} have the largest effects (with contributing value $-41.87$ and $-37.46$) that push the energy prediction lower from the base value to the predictive value. The other high-influence parameters are \texttt{shuffle.sort.bypassMergeThreshold}, \texttt{uncore.freq}, \texttt{hyperthreading}. For \texttt{bayes} in Figure~\ref{fig:shap_eup_bayes}, \texttt{uncore.freq} and \texttt{n.sockets} have the largest effects (with contributing value $-27.69$ and $-23.53$) that push the energy prediction lower from
the base value to the predictive value. The other high-influence parameters are \texttt{cpu.freq}, \texttt{n.cores}, \texttt{shuffle.file.buffer}, etc..

Two key points can be seen from these examples. First, both hardware
(e.g., \texttt{uncore.freq}) and software (e.g.,
\texttt{shuffle.sort.bypassMergeThreshold}) parameters influence the system
behavior, and thus both types of parameters should be tuned to meet
the optimization goal. Second, same parameters have different
effects on different workloads, which suggests that the optimal
configuration is workload dependent. Therefore, we should be careful
about the prior knowledge obtained from the existing workloads since
information learned for one workload might not transfer directly to
a new workload.

%% file: tex/conclusion.tex
\section{Limitations}\label{sec:limitation}

We recognize the limitations of this work as follows:
\begin{itemize}
	\item The outcome of predictive early termination and censored regression
	from \SYSTEM{} depends on the SEML framework. We chose to implement 
	\SYSTEM{} by adding these components to Bayesian optimization because 
	it has proven to provide high-quality samples. Future work can explore how \SYSTEM{} can be generalized to other learning frameworks.
	\item It is challenging to determine the search time budget in advance
	to make sure that the search results converge to the optimal. There
	has been work investigating this topic in the machine learning
	community~\cite{lorenz2015stopping,nguyen2017regret}.
	Although it is out of scope in this paper, it is interesting to
	explore the combination of \SYSTEM{} and this direction.
	\item The improper setting of \texttt{num\_boost\_round}, the parameter that controls the number of training steps can result in poor prediction accuracy. 
	Future work can explore more systematic ways to set this parameter, or incorporate its regularizing effect into the loss function.
\end{itemize}

\section{Conclusion} \label{sec:conclusion}

This paper presents \SYSTEM{}, a systematic approach to reduce the cost of samples through predictive early termination and censored regression. 
Evaluations show that \SYSTEM{} improves the system outcomes compared to a wide range of existing SEML and early termination techniques. Finally, we hope that \SYSTEM{} can inspire future research directions on reducing time cost of sample collection rather than the number of samples, and other applications of censored regression in computer systems research.